\documentclass[journal]{IEEEtran}
\usepackage{amsmath,amsfonts}
\usepackage{algorithm}
\usepackage{algorithmic}
\usepackage{array}
\usepackage[caption=false,font=normalsize,labelfont=sf,textfont=sf]{subfig}
\usepackage{textcomp}
\usepackage{stfloats}
\usepackage{url}
\usepackage{verbatim}
\usepackage{graphicx}
\usepackage{cite}
\usepackage{booktabs}
\usepackage{multirow}
\usepackage{newfloat}
\usepackage{listings}
\usepackage{makecell}

\begin{document}

\title{Mitigating Overthinking in Large Reasoning Models via Difficulty-aware Reinforcement Learning}
\author{Qian~Wan,
        Ziao~Xu,
        Luona~Wei,
        Xiaoxuan~Shen,
        and~Jianwen~Sun
\thanks{This work was financially supported by the National Key Research and Development Program of China under Grant 2024YFC3308200, the National Natural Science Foundation of China under Grant 62437002, Grant 62307015, Grant 62293554, and Grant 62577028, the Youth AI Talents Fund of the Chinese Association of Automation under Major Program under Grant HBRC-JKYZD-2024-310, the Hubei Provincial Natural Science Foundation of China under Grant 2024AFB169 and Grant 2023AFA020, and the Natural Science Foundation of Wuhan under Grant 2025040601020160. (\textit{Corresponding author}: Jianwen Sun.)}
\thanks{Q. Wan, X. Shen, and J. Sun are with the Laboratory for Artificial Intelligence and New Forms of Education, the National Engineering Research Center of Educational Big Data, and the Faculty of Artificial Intelligence in Education, Central China Normal University, Wuhan 430079, China (e-mail: wanq8228@ccnu.edu.cn; shenxiaoxuan@ccnu.edu.cn; sunjw@ccnu.edu.cn).}
\thanks{Z. Xu is with the National Engineering Research Center of Educational Big Data and the Faculty of Artificial Intelligence in Education, Central China Normal University, Wuhan 430079, China (e-mail: xuziao@mails.ccnu.edu.cn).}
\thanks{L. Wei is with the College of Electronics and Information Engineering, South-Central Minzu University, Wuhan 430074, China (e-mail: wlnelysion@scuec.edu.cn).}}

\markboth{Journal of \LaTeX\ Class Files,~Vol.~14, No.~8, August~2021}%
{Shell \MakeLowercase{\textit{et al.}}: A Sample Article Using IEEEtran.cls for IEEE Journals}


\maketitle

\begin{abstract}
Large Reasoning Models (LRMs) achieve explicit chain-of-thought expansion by imitating deep thinking behaviors of humans, demonstrating excellent performance in complex task scenarios. However, the deep-thinking mode often leads to unnecessarily lengthy reasoning and resource inefficiency when handling simple tasks. This overthinking phenomenon may arise from the generation preference triggered by the reward function during post-training. Existing research attempts to mitigate overthinking from the perspective of prompt design or model training, but generally underestimates the importance of task difficulty awareness, which makes it difficult for LRMs to effectively allocate reasoning resources. In this paper, we propose \textbf{Di}fficulty-aware \textbf{P}olicy \textbf{O}ptimization (DiPO), a reinforcement learning-based LRM training framework. DiPO encourages LRM to spontaneously model task complexity, and integrates them into reinforcement learning framework to adjust the generation preferences introduced by post-training. A difficulty modeling method based on model self-reasoning is proposed, which significantly reduces the dependence on manual annotation and formalize task complexity. We further develop a difficulty-signal-enhanced reward function that incorporates a penalty for lengthy reasoning while considering reasoning performance and output format. Experimental results indicate that DiPO enables the model to spontaneously adjust inference overhead, significantly reducing redundant tokens without losing performance due to thought compression.
\end{abstract}
\section{Introduction}
In recent years, large language models (LLMs) with deep thinking patterns, also known as large reasoning models (LRMs), have made significant strides in natural language processing \cite{openai2024openaio1card,deepseekai2025deepseekr1incentivizingreasoningcapability}. LRMs not only generate fluent text, but also demonstrate multi-step decomposition reasoning skills in tasks such as mathematical derivations, complex question answering, and planning. Generating longer chains-of-thought \cite{wei2023chainofthoughtpromptingelicitsreasoning} can significantly improve inference performance and interpretability of LRMs, especially for complex task scenarios that require step-by-step \cite{kojima2023largelanguagemodelszeroshot} decomposition. 

\begin{figure}
    \centering
    \includegraphics[width=1\linewidth]{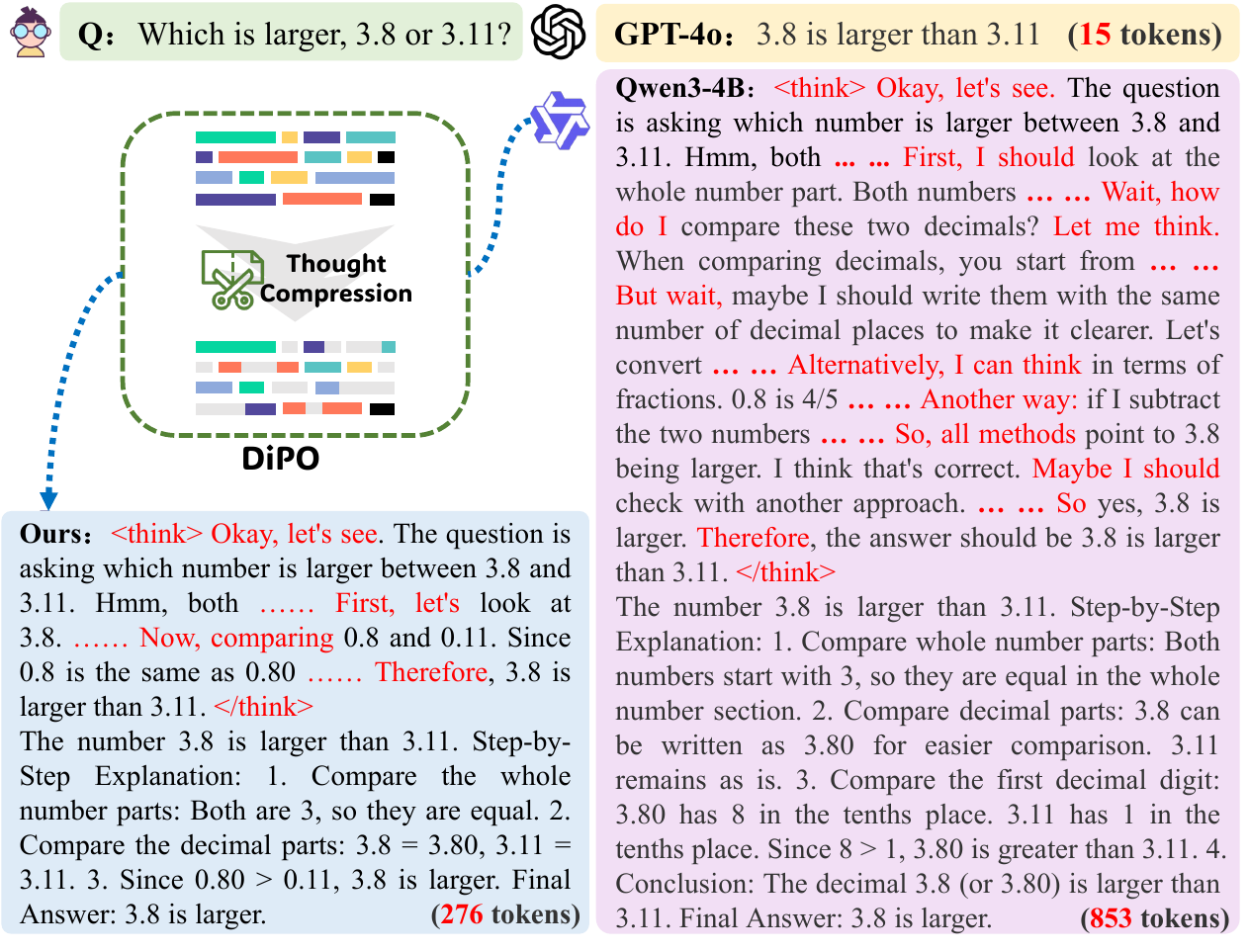}
    \caption{Responses from different models to the same simple question. This example was tested in June 2025.}
    \label{fig:fig1-introduction}
\end{figure}

Expanding the length of thought allows LRMs to simulate more detailed reasoning paths and iteratively verify answers, but it also introduces unnecessary redundant tokens and inference burden for simple tasks \cite{wang2022selfconsistency,han2025yourmodelsthoughtenough,fan2025missingpremiseoverthinking}. 
As shown in Figure \ref{fig:fig1-introduction}, when facing a simple question, "Which is larger, 3.8 or 3.11?", Qwen3-4B \cite{yang2025qwen3technicalreport} (a LRM capable of deep thought) generates 853 tokens, whereas the LLM GPT-4o \cite{openai2024gpt4technicalreport} uses only 13 tokens. 
Researchers have found that LRMs like OpenAI o1 \cite{openai2024openaio1card} often produce lengthy and repetitive responses for simple questions \cite{chen2025think23overthinkingo1like}. 
This phenomenon, known as overthinking, is particularly evident in scenarios that require quick responses or concise answers \cite{fan2025missingpremiseoverthinking}. 
Although the original design intention of LRMs is to enhance the inference performance of LLMs on complex tasks, the post-training stage inadvertently introduces unnecessary generation preferences for simple tasks, resulting in lengthy, repetitive model responses with unnecessary multiple verifications \cite{han2025yourmodelsthoughtenough,huang2025manifoldsteeringoverthinking}.

The root cause of overthinking may lie in the preference design of reward function during post-training process, specifically in fully rewarding correctness while neglecting the cost of generation.
The training objective is designed to provide correct answers within a fixed token length limit, encouraging the model to generate more tokens to enhance inference performance, which indirectly cultivates the habitual behavior of overthinking in the model. For this issue, prompt engineering might be a low-risk solution. This training-independent method not only avoids the potential loss of original performance but also significantly reduces computational burdens. For example, both CCoT \cite{renze2024benefits} and CoD \cite{xu2025chaindraftthinkingfaster} leverage manually designed prompts to extract the inherent capabilities of LRMs, aiming to improve inference efficiency without compromising performance. However, prompt-based methods heavily rely on human intervention, which lacks flexibility and strictly restricts the instruction-following ability of LRMs, thereby limiting its generalizability.

Examining the overthinking phenomenon from the perspective of model training may be closer to the root cause. Recent studies, on one hand, introduce length-coordination rewards during reinforcement learning process to reduce the model preference for generating lengthy sequences \cite{luo2025o1prunerlengthharmonizingfinetuningo1like,aggarwal2025l1controllinglongreasoning}. On the other hand, LRMs are trained using supervised data to skip low-priority tokens during inference \cite{xia2025tokenskipcontrollablechainofthoughtcompression,kang2024c3otgeneratingshorterchainofthought}, thereby selecting the shortest generation path. However, these methods fail to recognize the importance of fostering the model's spontaneous perception of task difficulty. The absence of difficulty signal modeling results in the inability to distinguish the complexity of different tasks, thereby allocating maximum computational resources in a generalized and monotonous manner during inference. The ability of difficulty-aware enables LRMs to effectively compress the reasoning process of simple tasks, while avoiding missing critical details due to thought compression when facing complex tasks. We believe that a controllable and explicit task difficulty perception system should be established for the model to alleviate overthinking, enabling it to dynamically adjust the reasoning depth according to the complexity of tasks.

In this study, we propose \textbf{Di}fficulty-aware \textbf{P}olicy \textbf{O}ptimization (DiPO), a reinforcement learning-based training framework, which aims at systematically mitigating the overthinking problem in LRMs. The core idea of DiPO is to cultivate the spontaneous perception of task difficulty in LRMs during inference process through post-training, and prompt them to assess complexity to reasonably compress lengthy reasoning steps. Specifically, we design a difficulty modeling method based on model self-reasoning. This method infers task complexity from the LRM output features, significantly reducing the dependence of the proposed training framework on manual annotation while formalizing the difficulty signal. We also introduce smoothing and normalization techniques in difficulty modeling process, effectively addressing the long-tail distribution issue exposed by the model output length and enhancing the modeling effectiveness. We further develop a difficulty-signal-enhanced reward function that incorporates a penalty for lengthy reasoning while considering reasoning performance and output format. The reward function is also implanted with a length-cropping mechanism to prevent extreme output lengths from negatively impacting the model. We use the prompt ``Let's think step by step’’ to maximize the reasoning capability and strictly constrain the output format. To our knowledge, this is the first attempt that formalizes task difficulty modeling to guide the compression thought of LRMs.

In the experimental section, we use training set and test sets from different sources to eliminate the possibility of data leakage. Experimental results on four intra-domain datasets show that the proposed method significantly reduces the token length while maintaining correctness, and the compression ratio of reasoning chain is significantly higher than baselines. Furthermore, the results on three out-of-domain datasets demonstrate that the training effect of proposed method possesses domain generalization capability.

Our contributions are summarized as follows:
\begin{itemize}

\item We propose a training framework based on reinforcement learning, named \textbf{Di}fficulty-aware \textbf{P}olicy \textbf{O}ptimization (DiPO), to alleviate the overthinking problem of LRMs.

\item A difficulty modeling method based on model self-reasoning is designed to formalize continuous difficulty signals, reducing the manual dependence of the proposed method.

\item We develop a difficulty-aware reward function that considers task complexity, balancing thought compression and accuracy metrics.

\item Extensive experiments demonstrate the effectiveness of DiPO, and relevant analysis can provide insights and inspiration for future research.

\end{itemize}

\section{Related Work}

\noindent{\textbf{Input Prompts-based Efficient Reasoning.}} These studies impose pre-reasoning constraints through prompt design or routing strategies to control reasoning length and complexity. Methods like Token-Budget \cite{hao2024traininglargelanguagemodels} set token limits, Chain-of-Draft (CoD) \cite{xu2025chaindraftthinkingfaster} restricts each step to a minimal draft, and Concise Chain-of-Thought (CCoT) \cite{renze2024benefits} uses prompts like ``Be concise’’ to simplify reasoning. In routing, RouteLLM \cite{ong2025routellmlearningroutellms} assigns tasks based on difficulty, Self-REF \cite{chuang2025learningroutellmsconfidence} directs uncertain tasks to stronger models, and Sketch-of-Thought (SoT) \cite{aytes2025sketchofthoughtefficientllmreasoning} adapts reasoning paradigms based on question features. These methods rely on manually designed templates and heuristic approaches, and the high demand for instruction compliance limits their scenario generalization.

\noindent{\textbf{Reasoning Output-based Efficient Reasoning.}} These works focus on controlling the compression or decoding of latent representations to enhance the ability of LRMs to reason concisely and efficiently. Methods like Coconut \cite{hao2024traininglargelanguagemodels} replace text generation with compressed hidden states, enabling more compact reasoning. CCoT \cite{cheng2024compressedchainthoughtefficient} precomputes key hidden states and uses LoRA \cite{hu2021loralowrankadaptationlarge} to shorten long chains into thought tokens. SoftCoT \cite{xu2025softcotsoftchainofthoughtefficient} generates soft thought tokens via a lightweight model and injects them into the embedding space for implicit reasoning. These technologies rely on posterior signals rather than model self-regulation to reduce lengthy outputs, and the introduction of additional strategies increases uncontrollable risks.

\noindent{\textbf{Model-based Efficient Reasoning.}} These studies focus on designing rewards to fine-tuning LRMs to improve their intrinsic ability to reason concisely and efficiently. For example, O1-Pruner \cite{luo2025o1prunerlengthharmonizingfinetuningo1like} uses a length-coordination reward with PPO \cite{schulman2017proximalpolicyoptimizationalgorithms} to reduce reasoning length while maintaining accuracy. Demystifying \cite{yeo2025demystifyinglongchainofthoughtreasoning} employs a cosine-based reward to manage performance fluctuations caused by excessive reasoning length, while L1 \cite{aggarwal2025l1controllinglongreasoning} adds explicit instructions for controlled reasoning length. In supervised fine-tuning with variable-length CoT data, methods like Self-Training \cite{munkhbat2025selftrainingelicitsconcisereasoning} select the shortest paths, TokenSkip \cite{xia2025tokenskipcontrollablechainofthoughtcompression} skips less important tokens, and C3oT \cite{kang2024c3otgeneratingshorterchainofthought} compresses redundancy using GPT-4. 

The above methods attempt to change LRMs generation preferences through fine-tuning, but none of them attach importance to cultivating the model ability to perceive task difficulty. Our study proposes a novel and efficient post-training method by explicitly modeling task difficulty, enabling the model to dynamically adjust the reasoning depth according to the task complexity.

\section{Method}

In this section, we propose DiPO, a reinforcement learning framework that dynamically adjusts reasoning depth based on task difficulty to address excessive reasoning for simple tasks and brief outputs for complex ones. The core idea is to use the model's output length as an implicit signal of task complexity, leveraging reinforcement learning for fine-tuning the model to adaptively control the reasoning depth. The method consists of several stages: difficulty signal definition, data reconstruction, reward function design, and reinforcement learning fine-tuning.

\noindent{\textbf{Task Definition.}} LRMs often exhibit a mismatch between reasoning length and task difficulty—simple tasks are overthought, while complex tasks receive insufficient reasoning. We formalize this as the problem of \textit{reasoning length misalignment}.
Given a task set $\mathcal{D} = \{(q_i, a_i)\}_{i=1}^N$, where $q_i$ is a natural language question and $a_i$ the reference answer, the goal is to train a model $\pi_\theta$ that maintains high accuracy while adjusting its reasoning length according to task complexity, balancing efficiency and performance.


\noindent{\textbf{Difficulty Signal Construction.}} 
To estimate task complexity without manual annotations, we propose an automated difficulty signal based on output length. For each training instance, the input \( q_i \) is transformed using a fixed prompting template, which adds the phrase ``let's think step by step'' to stimulate the model's reasoning process and instructs the model to output the final answer within a \(\texttt{\textbackslash box\{\}}\). Subsequently, we employ a large language model with strong mathematical problem-solving and reasoning capabilities to perform generative reasoning, producing the model's output text answer \( y_i \).
In order to provide meaningful feedback signals for the reinforcement learning reward function, we consider not only the length of the response but also the correctness of the output.
We define the response correctness as a binary indicator $\delta_i$:
\begin{equation}
\delta_i =
\begin{cases}
0, & \text{if } y_i \text{ is correct}, \\
1, & \text{if } y_i \text{ is not correct}.
\end{cases}
\end{equation}

Let \( L(x_i) = \text{len}(y_i) \) denote the number of tokens in the generated response. However, directly using $L(x_i)$ as a reward may face long-tail distribution problems, where excessively long sequences result in large fluctuations in the reward signal, which can lead to training instability. To address this, we apply a smoothing transformation to the length to reduce the impact of extreme values on training. For detailed information, please refer to experimental section. 
\begin{equation}
\tilde{L}(x_i) = \sqrt{L(x_i)}
\end{equation}

To ensure consistency and comparability of the difficulty signal across all samples, we further standardize the smoothed length. Let the mean and standard deviation of all $\tilde{L}(x_i)$ in the training set be $\mu$ and $\sigma$, respectively. The standardized difficulty score is then defined as:
\begin{equation}
Z(x_i) = \frac{\tilde{L}(x_i) - \mu}{\sigma} + \alpha \cdot \delta_i
\end{equation}
where $\mu$ and $\sigma$ can be pre-computed over the entire training set or dynamically estimated using a sliding window. In this paper, we choose the pre-computed method to obtain the mean and standard deviation. The hyperparameter $\alpha$ is the error penalty coefficient, used to control the contribution of answer correctness to the final difficulty signal, thus balancing the impact of answer correctness and response length in the reward design.

Before outputting the final difficulty signal, we apply the clipping function to process the difficulty signal in order to mitigate the impact of extreme values on the experiment.
\begin{equation}
\text{Diff}(x_i) = \text{clip}(Z(x_i), 1 - \xi, 1 + \xi)
\end{equation}
where \(\xi\) is a hyperparameter that controls the clipping function.

\noindent{\textbf{Difficulty-Annotated Dataset.}}
We augment the original dataset with the difficulty scores:
\begin{equation}
\mathcal{D}^{\text{new}} = \{(x_i, a_i, \text{Diff}(x_i))\}_{i=1}^N
\end{equation}

This enriched dataset provides reward-aware supervision for downstream reinforcement learning.

\noindent{\textbf{Reward Function Design.}}
During the reinforcement learning training process, the LLM receives a sequence \( x = [x_1, \dots, x_n] \) as input and generates a corresponding solution \( o = [o_1, \dots, o_n] \), with \( \hat{a}_i \) representing the extracted answer form \( o_i \). The reward function penalizes unnecessary length while maintaining correctness:
\begin{equation}
r_i(a_i, \hat{a}_i, \text{len}(o_i), \text{Diff}(q_i)) =
\begin{cases}
p, & \hat{a}_i = \text{none} \\
f - \lambda_i, & \hat{a}_i \ne a_i \\
s - \lambda_i, & \hat{a}_i = a_i \\
\end{cases}
\end{equation}
The definition of $\lambda_i$ is as follows:
\begin{equation}
\lambda_i = \min\left(\epsilon, \frac{\text{len}(o_i)}{c}\right) \cdot (\text{Diff}(q_i) + \varphi)
\end{equation}

This design rewards correctness (score \( s \)) and penalizes both formatting errors (score \( p \)) and incorrect answers (score \( f \)), scaled by response length and task difficulty. The parameters \( c \), \( \epsilon \), and \( \varphi \) regulate the range of penalty values and stabilize learning.

\noindent{\textbf{Reinforcement Learning Procedure.}}
After defining the reward, we proceed to optimize the model using GRPO \cite{shao2024deepseekmathpushinglimitsmathematical}. Our method can be seen as a difficulty-aware variant of GRPO, incorporating task complexity into the reward to guide policy updates.

Let $\pi_\theta$ denote the policy network (i.e. LRM) and $\pi_{\text{ref}}$ the frozen reference model. For each input $q_i$, we generate $K$ candidate outputs $\{o_i^k\}_{k=1}^K$ from the current policy, compute their rewards $\{r_i^k\}_{k=1}^K$, and apply a weighted policy gradient to update $\theta$. The loss function is:

\begin{equation}
\mathcal{L}^{\text{RL}}(\theta) = - \sum_{i=1}^N \sum_{k=1}^K w_i^k \log \pi_\theta(o_i^k \mid q_i)
\end{equation}
where the importance weights are defined as:

\begin{equation}
w_i^k = \frac{\exp(\beta (r_i^k - b_i))}{\sum_{j=1}^K \exp(\beta (r_i^j - b_i))}
\end{equation}
with baseline $b_i = \frac{1}{K} \sum_{k=1}^K r_i^k$ representing the average reward of the current batch, and $\beta > 0$ controlling the sharpness of the reward-based weighting. This formulation ensures that candidates with higher relative rewards receive proportionally larger gradients, effectively focusing the optimization on promising outputs while maintaining training stability.To the end, we summarize the training procedure of our proposed DiPO in Algorithm \ref{alg:DiPO}.

\begin{algorithm}[tb]
\caption{Difficulty-aware Policy Optimization}
\label{alg:DiPO}
\begin{flushleft}
\textbf{Input:} LLM \( \pi_\theta \), dataset \( \mathcal{D} = \{(q_i, a_i, \text{Diff}(q_i))\}_{i=1}^N \) \\
\textbf{Output:} Updated LLM \( \pi_{\theta'} \)
\end{flushleft}
\begin{algorithmic}[1]
\STATE Initialize reference model \( \pi_{\text{ref}} \leftarrow \pi_\theta \)
\FOR{$i = 1$ \TO $N$}
    \STATE Sample $K$ solutions $o_i^1, \ldots, o_i^K$ from $\pi_{\text{ref}}(\cdot \mid q_i)$
    \FOR{$k = 1$ \TO $K$}
        \STATE Extract answer $\hat{a}_i^k$ from $o_i^k$
        \STATE Compute correctness: $\delta_i^k = I[\hat{a}_i^k \neq a_i]$
        \STATE Compute output length: $L(o_i^k) = \text{len}(o_i^k)$
        \STATE Compute difficulty penalty: $\lambda_i^k$ 
        \STATE Assign reward: $r_i^k$
    \ENDFOR
    \STATE Compute GRPO loss $\mathcal{L}^{\text{RL}}(\theta)$ using $\{r_i^k\}_{k=1}^K$
    \STATE Update parameters: $\theta' \leftarrow \theta - \eta \nabla_\theta \mathcal{L}^{\text{RL}}$
\ENDFOR
\end{algorithmic}
\end{algorithm}

\noindent{\textbf{Summary.}} Our method introduces a unified pipeline that quantifies task difficulty using model-internal signals and incorporates it into reinforcement learning to adaptively control the depth of reasoning. The result is a flexible, automated optimization framework that mitigates overthinking and enhances reasoning efficiency across diverse tasks.

\section{Experimental Settings}

\subsection{Datasets}
We train on the TAL-SCQ5K dataset \cite{TAL-SCQ}, which includes 5,000 math questions in English and Chinese. We use the English version and convert it into fill-in-the-blank questions with DeepSeek-V3, making it suitable for Chain-of-Thought training. We evaluate DiPO on several in-domain and out-of-domain datasets. For in-domain datasets, we use GSM8K \cite{cobbe2021training}, Math-500 \cite{NEURIPSDATASETSANDBENCHMARKS2021_be83ab3e,lightman2023letsverifystepstep}, GaoKao \cite{zhang2024evaluatingperformancelargelanguage}, and AIME 2025, all of which focus on mathematical reasoning tasks. For out-of-domain datasets, we evaluate on GPQA \cite{rein2023gpqagraduatelevelgoogleproofqa}, LAST \cite{zhong2023agievalhumancentricbenchmarkevaluating}, and MMLU \cite{hendrycks2021measuringmassivemultitasklanguage}, which cover a broad range of tasks across different domains. It is worth noting that we use training set and test sets from different sources to eliminate the possibility of data leakage.

\subsection{Baseline}
We evaluate our method by comparing it with the following representative approaches aimed at enhancing reasoning efficiency:

\begin{itemize}
    \item \textbf{TALE-EP} \cite{han2025tokenbudgetawarellmreasoning}: A prompt-based approach that estimates a token budget before reasoning to control inference length and cost.
    
    \item \textbf{Chain of Draft (CoD)} \cite{xu2025chaindraftthinkingfaster}: This is another prompt-based method which instructs the model to generate concise draft intermediate steps during reasoning.
    
    \item \textbf{SFT} \cite{munkhbat2025selftrainingelicitsconcisereasoning}: This approach selects the shortest correct outputs from sampled responses for supervised fine-tuning(SFT). We use the two shortest among 8 correct samples.
    
    \item \textbf{DPO} \cite{rafailov2024directpreferenceoptimizationlanguage}: This method fine-tunes preferences using pairs of the longest, shortest, and second shortest correct responses, encouraging more efficient reasoning by optimizing for shorter answers.
\end{itemize}

\subsection{Evaluation Metrics}
We use the following metrics to jointly evaluate reasoning accuracy and compression efficiency:
\begin{itemize}
    \item \textbf{ACC}: Denotes the accuracy of the final answer.
    \item \textbf{LEN}: Refers to the average response length, measured in tokens.
    \item \textbf{Ratio}: Represents the proportion of the token position where the correct answer first appears to the total response length in the current output \cite{chen2025think23overthinkingo1like}.
\end{itemize}

\begin{table*}[t]
\centering
\caption{Results of applying different optimization strategies to Qwen3-4B and DeepSeek-R1-0528-Qwen3-8B (denoted as Qwen3-8B$^{*}$) on four math-reasoning benchmarks (GSM8K, Math-500, GaoKao, AIME-2025). LEN reports the average response length (tokens). The percentage in parentheses denotes the length normalized by the corresponding Baseline within each model and dataset. Ratio is the average token-saving ratio (higher is better).}
\label{table:all_results}
\setlength{\tabcolsep}{4pt} 
\begin{tabular}{lcccccccccccc}
\toprule
\multirow{2}{*}[-0.5ex]{Model} & \multicolumn{3}{c}{GSM8K} & \multicolumn{3}{c}{Math-500} & \multicolumn{3}{c}{GaoKao} & \multicolumn{3}{c}{AIME-2025} \\
\cmidrule(lr){2-4} \cmidrule(lr){5-7} \cmidrule(lr){8-10} \cmidrule(lr){11-13}
& ACC & LEN & Ratio & ACC & LEN & Ratio & ACC & LEN & Ratio & ACC & LEN & Ratio \\
\midrule
\textit{Qwen3-4B} & & & & & & & & & & & & \\
Baseline & \textbf{94.62} & \makecell{1097.4 \\ (100.00\%)} & 0.415 & 87.8 & \makecell{3187.2 \\ (100.00\%)} & 0.311 & 75.58 & \makecell{3208.1 \\ (100.00\%)} & 0.402 & 26.67 & \makecell{7441.9 \\ (100.00\%)} & 0.416 \\
DPO & 94.09 & \makecell{1087.7 \\ (99.12\%)} & 0.410 & 88.8 & \makecell{3181.0 \\ (99.81\%)} & 0.313 & 77.14 & \makecell{3180.7 \\ (99.15\%)} & 0.403 & 26.67 & \makecell{7572.4 \\ (101.75\%)} & 0.434 \\
SFT & 94.24 & \makecell{1086.2 \\ (98.98\%)} & 0.412 & 90.0 & \makecell{3185.3 \\ (99.94\%)} & 0.318 & 75.84 & \makecell{3252.0 \\ (101.37\%)} & 0.400 & 26.67 & \makecell{7427.3 \\ (99.80\%)} & 0.425 \\
TALE-EP & 94.39 & \makecell{561.5 \\ (51.17\%)} & 0.497 & 91.8 & \makecell{2046.1 \\ (64.19\%)} & 0.388 & \textbf{79.74} & \makecell{2139.7 \\ (66.69\%)} & 0.493 & 40.00 & \makecell{6883.5 \\ (92.50\%)} & 0.654 \\
CoD & 94.16 & \makecell{464.2 \\ (42.30\%)} & 0.558 & 90.0 & \makecell{1786.4 \\ (56.05\%)} & 0.474 & 79.14 & \makecell{1799.1 \\ (56.08\%)} & 0.578 & 36.67 & \makecell{6562.6 \\ (88.18\%)} & 0.678 \\
DiPO & 94.31 & \makecell{\textbf{362.5} \\ \textbf{(33.03\%)}} & \textbf{0.582} & \textbf{92.20} & \makecell{\textbf{1354.8} \\ \textbf{(42.51\%)}} & \textbf{0.513} & \textbf{79.74} & \makecell{\textbf{1490.8} \\ \textbf{(46.47\%)}} & \textbf{0.593} & \textbf{43.33} & \makecell{\textbf{5950.1} \\ \textbf{(79.95\%)}} & \textbf{0.777} \\
\midrule
\textit{Qwen3-8B}$^{*}$ & & & & & & & & & & & & \\
Baseline & 93.93 & \makecell{1726.3 \\ (100.00\%)} & 0.267 & 81.60 & \makecell{3895.0 \\ (100.00\%)} & 0.228 & 71.43 & \makecell{3976.4 \\ (100.00\%)} & 0.279 & 16.67 & \makecell{7633.0 \\ (100.00\%)} & 0.425 \\
DPO & 93.63 & \makecell{1740.1 \\ (100.80\%)} & 0.267 & 82.60 & \makecell{3898.3 \\ (100.08\%)} & 0.246 & 71.17 & \makecell{3952.7 \\ (99.40\%)} & 0.285 & 16.67 & \makecell{7755.8 \\ (101.61\%)} & 0.375 \\
SFT & 94.09 & \makecell{1448.1 \\ (83.89\%)} & 0.296 & 83.20 & \makecell{3638.3 \\ (93.41\%)} & 0.266 & 74.29 & \makecell{3655.3 \\ (91.92\%)} & 0.295 & 20.00 & \makecell{7552.7 \\ (98.95\%)} & 0.433 \\
TALE-EP & 93.33 & \makecell{1846.0 \\ (106.93\%)} & 0.205 & 83.20 & \makecell{3772.9 \\ (96.86\%)} & 0.238 & 71.69 & \makecell{3775.6 \\ (94.95\%)} & 0.260 & 16.67 & \makecell{7573.2 \\ (99.22\%)} & 0.442 \\
CoD & \textbf{94.16} & \makecell{2123.4 \\ (122.99\%)} & 0.187 & 82.80 & \makecell{3781.5 \\ (97.09\%)} & 0.217 & 72.99 & \makecell{3799.0 \\ (95.54\%)} & 0.251 & 20.00 & \makecell{7545.1 \\ (98.85\%)} & 0.435 \\
DiPO & 93.78 & \makecell{\textbf{494.7} \\ \textbf{(28.66\%)}} & \textbf{0.630} & \textbf{83.40} & \makecell{\textbf{1360.1} \\ \textbf{(34.92\%)}} & \textbf{0.567} & \textbf{74.55} & \makecell{\textbf{1575.2} \\ \textbf{(39.62\%)}} & \textbf{0.595} & \textbf{30.00} & \makecell{\textbf{6157.5} \\ \textbf{(80.67\%)}} & \textbf{0.701} \\
\bottomrule
\end{tabular}
\end{table*}

\subsection{Implementation Details}
During the construction of $\mathcal{D}^{\text{new}}$, we precompute the mean $\mu$ and standard deviation $\sigma$ of $\tilde{L}(x_i)$, set the error penalty coefficient $\alpha$ to 0.1, the difficulty bias $\varphi$ to 0.8, and the clipping function parameter $\xi$ to 0.8. For GRPO \cite{shao2024deepseekmathpushinglimitsmathematical} training, we set a small learning rate of 3e-7 and use a batch size of 64 to ensure the stability and efficiency of the training process. To minimize the risk of answer truncation, we set the maximum context length to 8K tokens.Training is conducted for 60 steps using the VeRL framework \cite{Sheng_2025}, and we generate 16 candidate responses per input prompt (n=16) to enhance data diversity and improve the stability of policy learning. In Section 5.8, we will explain why the number of training steps was set to 60. During training, we assign a base score $s$ of 1.0 for correct answers, a format score $f$ of -0.2 for responses with correct format but wrong answers, and the format penalty $p$ is set to -1 when the answer cannot be extracted from the response. For the length constraint, we set the length scaling factor $c$ to 9000 and the maximum length penalty threshold $\epsilon$ to 0.5.

We conducted experiments on two LRMs, Qwen3-4B \cite{yang2025qwen3technicalreport} and DeepSeek-R1-0528-Qwen3-8B \cite{deepseek2025R1-0528}. This setup allows exploration of the effectiveness of DiPO on models at different training stages, namely post-training and data distillation. Qwen3-4B is treated as a pure reasoning model in our experiments, as we disregard its official settings that enforce a restriction on reasoning length to avoid introducing potential bias. \textbf{To ensure fairness, all methods and baselines were evaluated using unified inference settings, specifically maintaining a constant temperature and a maximum context length of 8K tokens across all runs.} All models are trained on a 4*A800-80G server. Training Qwen3-4B takes approximately 36 hours, while DeepSeek-R1-0528-Qwen3-8B requires 72 hours.

\begin{table*}[t]
\centering
\caption{Performance evaluation of models on OOD datasets, including GPQA, MMLU, and LAST. The DiPO achieves the highest accuracy and ratio across all datasets, demonstrating superior performance compared to other methods.}
\label{table:ooddataset}
\setlength{\tabcolsep}{5pt}
\begin{tabular}{lcccccccccc}
\toprule 
\multirow{2}{*}[-0.5ex]{Model} & \multicolumn{3}{c}{GPQA} & \multicolumn{3}{c}{MMLU} & \multicolumn{3}{c}{LAST} \\
\cmidrule(lr){2-4} \cmidrule(lr){5-7} \cmidrule(lr){8-10}
 & ACC & LEN & Ratio & ACC & LEN & Ratio & ACC & LEN & Ratio \\
\midrule 
\textit{Qwen3-4B} & & & & & & & & & \\
Baseline & 35.71 & 5452.9 (100.00\%) & 0.737 & 78.97 & 1365.8 (100.00\%) & 0.852 & 67.83 & 5250.8 (100.00\%) & 0.617 \\
DPO & 35.94 & 5349.1 (98.10\%) & 0.752 & 78.64 & 1381.0 (101.11\%) & 0.853 & 71.30 & 5205.9 (99.15\%) & 0.658 \\
SFT & 36.16 & 5370.8 (98.50\%) & 0.777 & \textbf{79.20} & 1338.5 (97.90\%) & 0.853 & 69.57 & 5181.8 (98.69\%) & 0.656 \\
TALE-EP & 41.52 & 3572.7 (65.52\%) & 0.791 & 77.70 & 744.1 (54.48\%) & 0.909 & 64.35 & 4415.9 (84.10\%) & 0.716 \\
CoD & 39.96 & 3060.6 (56.13\%) & 0.834 & 75.54 & 777.6 (56.93\%) & \textbf{0.929} & 70.00 & 3828.6 (72.92\%) & 0.745 \\
DiPO & \textbf{43.53} & \textbf{2769.0} (50.78\%) & \textbf{0.857} & 78.79 & \textbf{534.3} (39.12\%) & 0.923 & \textbf{72.17} & \textbf{2979.1} (56.74\%) & \textbf{0.767} \\
\bottomrule
\end{tabular}
\end{table*}

\section{Results and Analysis}

\subsection{Overall Performance}
Table \ref{table:all_results} presents the performance of various methods across different evaluation metrics. Compared to baseline and competitive methods, DiPO consistently strikes a superior balance between reasoning accuracy and efficiency. Notably, DiPO achieves the best trade-off between accuracy and reasoning length, with a significantly reduced reasoning length (LEN) and an accuracy (ACC) either close to or higher than the baseline. In experiments on the Math-500 dataset, DiPO achieves 92.2\% accuracy with a 57.4\% reduction in solution length for Qwen3-4B, and a 65.0\% reduction for DeepSeek-R1-0528-Qwen3-8B, while improving accuracy. These results demonstrate DiPO's robustness in enhancing reasoning efficiency without sacrificing accuracy.

In comparison to prompt-based methods such as TALE-EP and CoD, DiPO shows better generalization. TALE-EP and CoD's efficiency improvements are less pronounced on DeepSeek-R1-0528-Qwen3-8B than on Qwen3-4B, suggesting that prompt-based methods rely heavily on model adherence to instructions. DiPO avoids this limitation. Additionally, methods like SFT and DPO are more sensitive to dataset quality.

Overall, DiPO requires less stringent data quality and is less dependent on the model's instruction-following ability, demonstrating superior and robust performance across datasets.

\subsection{Analysis of Out-of-Domain Datasets}

\begin{figure}
    \centering
    \includegraphics[width=1\linewidth]{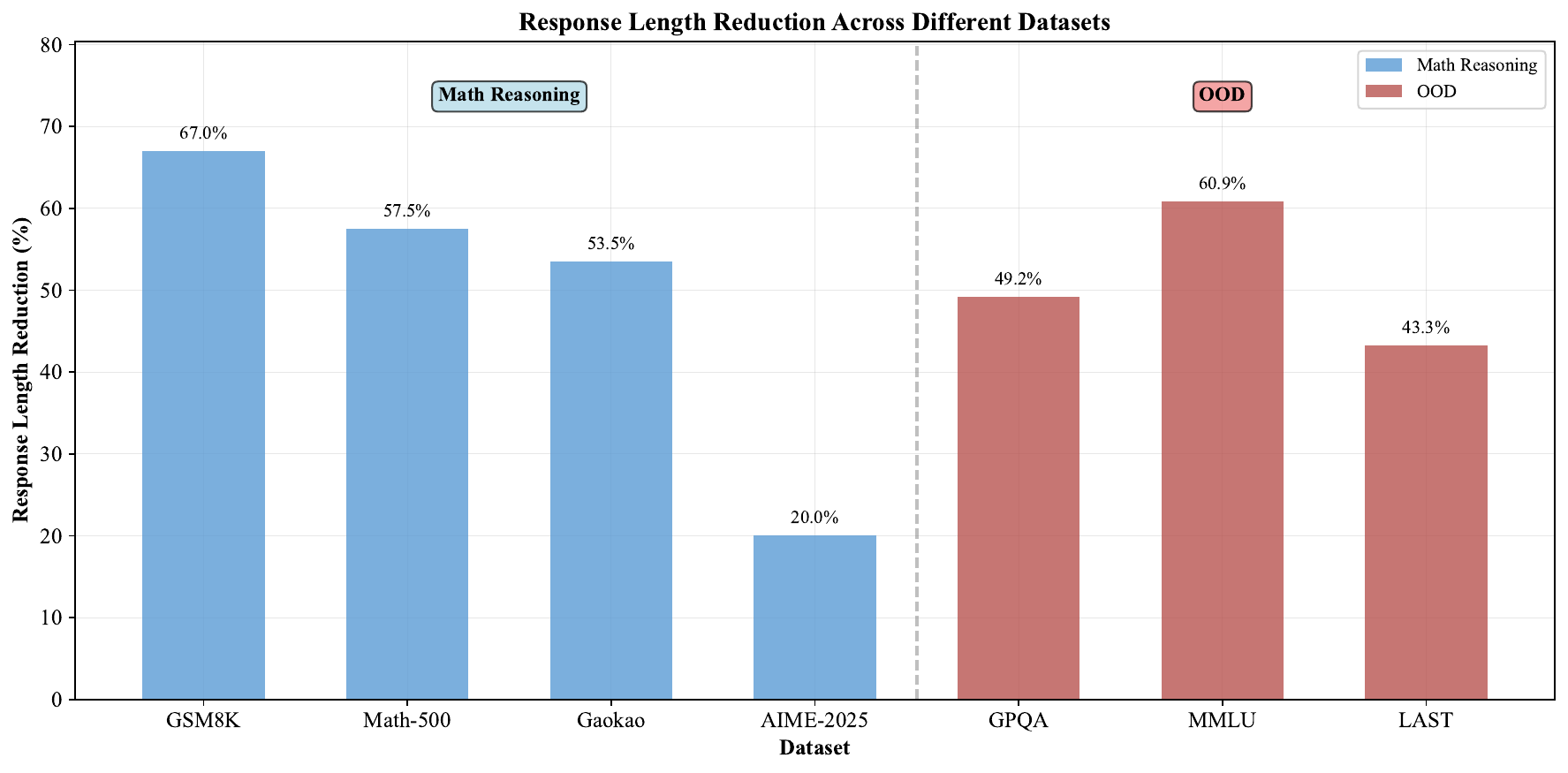}
    \caption{\textbf{Response Length Reduction Across Different Datasets.} Response length reduction across different datasets, with the y-axis representing percentage token reduction for math reasoning (left) and OOD datasets (right).}
    \label{fig:Inference-Compression-Performance-Analysis}
\end{figure}

DiPO demonstrates effective generalization to out-of-domain (OOD) tasks. We evaluate its ability to transfer reasoning compression to non-mathematical reasoning domains by using Qwen3-4B as the base model and conducting additional experiments on three out-of-domain datasets: GPQA, MMLU, and LAST. Table \ref{table:ooddataset} presents the results. Compared to other methods, DiPO exhibits remarkable control over reasoning length across nearly all datasets. For example, in the MMLU dataset, DiPO reduces reasoning length by 60.9\% compared to the baseline while maintaining comparable accuracy. In the more challenging GPQA dataset, DiPO not only shows a significant improvement in reasoning efficiency but also achieves a 7.82\% increase in accuracy compared to the baseline, with a notable improvement in the Ratio as well. These findings indicate that DiPO excels not only on standard datasets but also demonstrates stable reasoning efficiency in out-of-domain tasks, effectively reducing redundant steps in the reasoning process while maintaining or even enhancing the quality of the model's output.

\subsection{Inference Compression Performance Analysis}

To explore the reasoning compression capability of DiPO across various datasets, we conducted experiments on multiple datasets. As shown in Figure \ref{fig:Inference-Compression-Performance-Analysis}, DiPO significantly reduces response lengths across seven different datasets, achieving compression rates ranging from 20.0\% to 67.0\%. In mathematical reasoning tasks, the method demonstrates a clear task complexity-adaptive feature: on the relatively simple GSM8K and Math-500 datasets, it achieves compression rates of 67.0\% and 57.5\%, respectively, while on the more challenging AIME-2025 dataset, the compression rate is 20.0\%. This suggests that the model can intelligently adjust the generated reasoning length based on task difficulty. This differentiated compression strategy effectively addresses the ``overthinking" phenomenon in simpler problems while retaining sufficient reasoning steps for more complex problems.

In terms of out-of-domain generalization, the method also performs excellently. It achieves compression rates of 49.2\% and 60.9\% on the GPQA and MMLU datasets, respectively, and 44.7\% on the LAST dataset. These results demonstrate the method's strong adaptability when handling different knowledge domains and reasoning styles. Even when facing task types that were not seen during training, the model can still effectively identify and compress redundant reasoning steps, maintaining an efficient generation strategy.

In summary, the analysis reveals that our method not only excels in specialized tasks such as mathematical reasoning but also exhibits strong robustness and efficiency across a diverse range of out-of-domain datasets, proving its outstanding task adaptability and cross-domain generalization ability.

\begin{figure}
    \centering
    \includegraphics[width=1\linewidth]{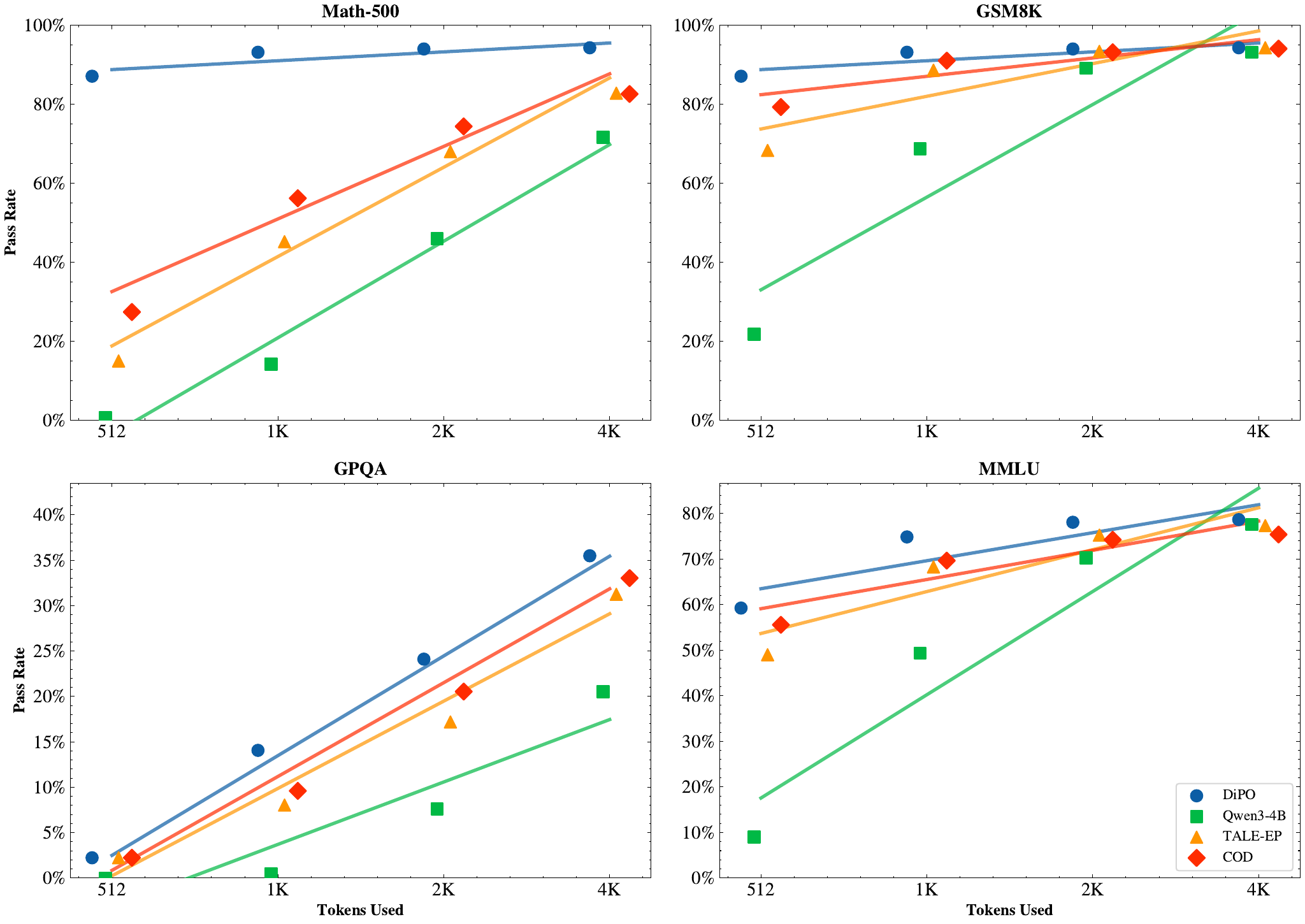}
    \caption{\textbf{Performance of DiPO with Different Token Budgets.}
The pass rate (y-axis) across various benchmarks is shown as a function of the number of tokens used (x-axis). }
    \label{fig:Inference-Efficiency Analysis}
\end{figure}

\subsection{Inference Efficiency Analysis}

Figure \ref{fig:Inference-Efficiency Analysis} illustrates the performance of DiPO under length constraints across different datasets, aiming to explore its performance when faced with length limitations. DiPO significantly improves the model's reasoning efficiency by reducing redundant reasoning. A comparison under the same token budget reveals its clear advantage in both reasoning compression efficiency and generalization ability. For instance, on the GSM8K and Math-500 datasets, when the token budget is limited to 512, DiPO achieves an accuracy of over 80\%, while the base model achieves less than 40\%, and in the Math-500 dataset, it even drops to 0. Compared to other methods, DiPO also demonstrates a significant improvement in reasoning efficiency. 
In handling the more complex GPQA dataset, accuracy slightly declines under lower token limits. This indicates that, for more difficult problems, the model requires more reasoning steps to arrive at the correct answer. It also highlights the flexibility of DiPO. The model can adaptively adjust the token length based on the complexity of the problem, meaning that it does not lose its original reasoning capability in pursuit of a shorter reasoning process. This further validates the model's adaptive ability to dynamically adjust the output length based on task difficulty. It mitigates redundant reasoning for simple tasks while retaining the depth needed for complex tasks.

\subsection{Performance Comparison Across Difficulty Levels}

\begin{figure}
    \centering
    \includegraphics[width=1\linewidth]{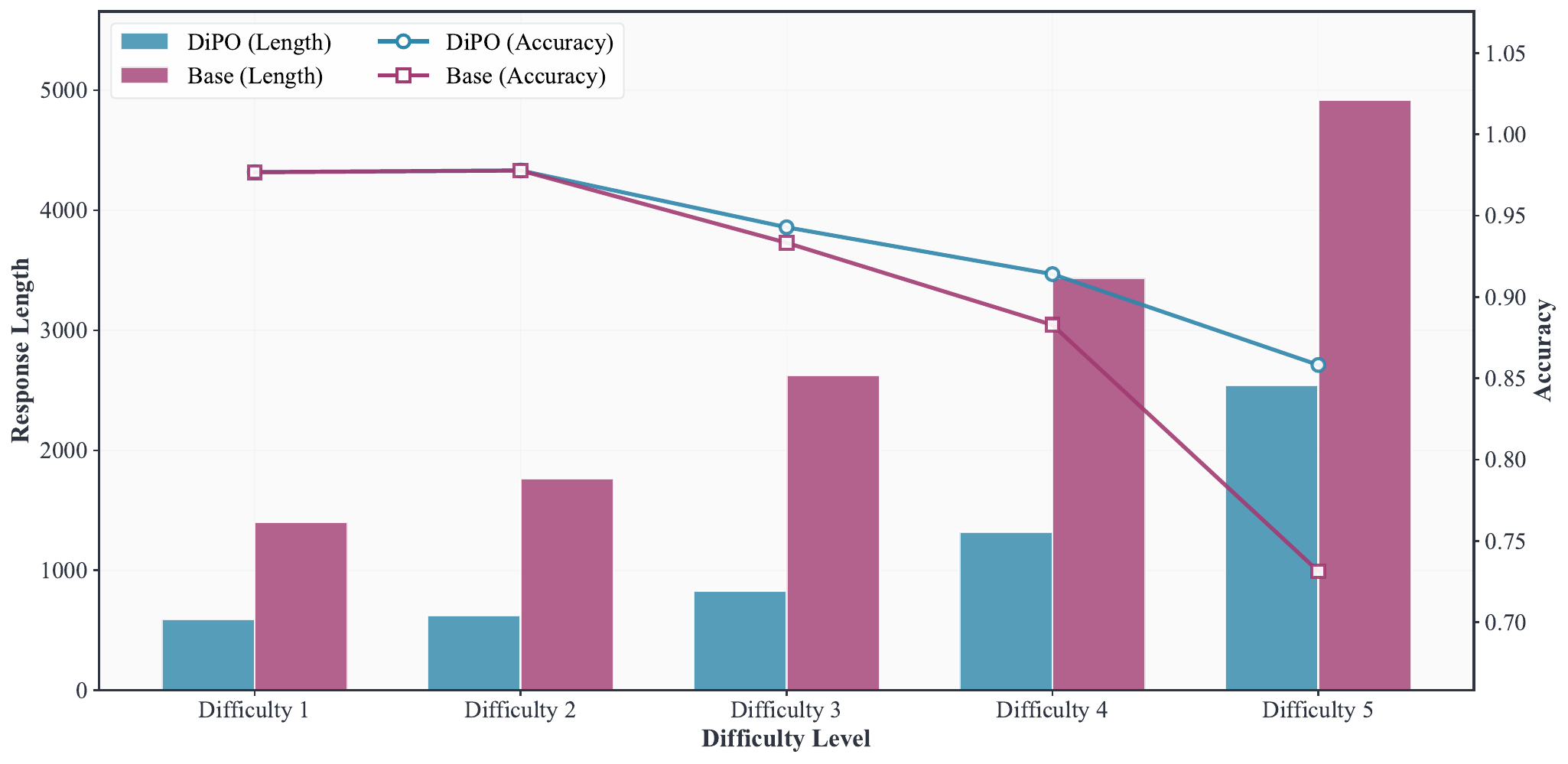}
    \caption{\textbf{Multi-model performance comparison across difficulty levels on the Math-500 dataset.} The x-axis represents the difficulty level of the tasks, and the bars correspond to the response lengths of the Qwen3-4B model (Base) and the DiPO, with accuracy shown by the lines for both models.}
    \label{fig:Performance-Comparison-Across-Difficulty-Levels}
\end{figure}

Figure \ref{fig:Performance-Comparison-Across-Difficulty-Levels} illustrates the significant advantages of DiPO across different difficulty levels, particularly in balancing accuracy and generation length. Whether handling simple tasks (e.g., Difficulty 1) or complex tasks (e.g., Difficulty 5), our model maintains high accuracy while generating relatively short outputs. At all difficulty levels, DiPO achieves over 50\% savings in computational resources, highlighting the superiority of the difficulty-aware adaptive mechanism. Moreover, we were surprised to find that DiPO seems to enhance the base model’s ability to solve more complex problems. For instance, on Difficulty 1 tasks, DiPO's accuracy matches that of the base model, but as task difficulty increases, the accuracy gap between DiPO and the base model widens. This indicates that DiPO not only improves reasoning efficiency but also enhances the base model's capacity to address more complex problems.

Overall, the experimental results demonstrate that, compared to the Qwen3-4B base model, DiPO shows significant advantages in reasoning compression, accuracy, and task adaptability, while also improving the model’s ability to solve complex problems.
\begin{figure}
    \centering
    \includegraphics[width=1\linewidth]{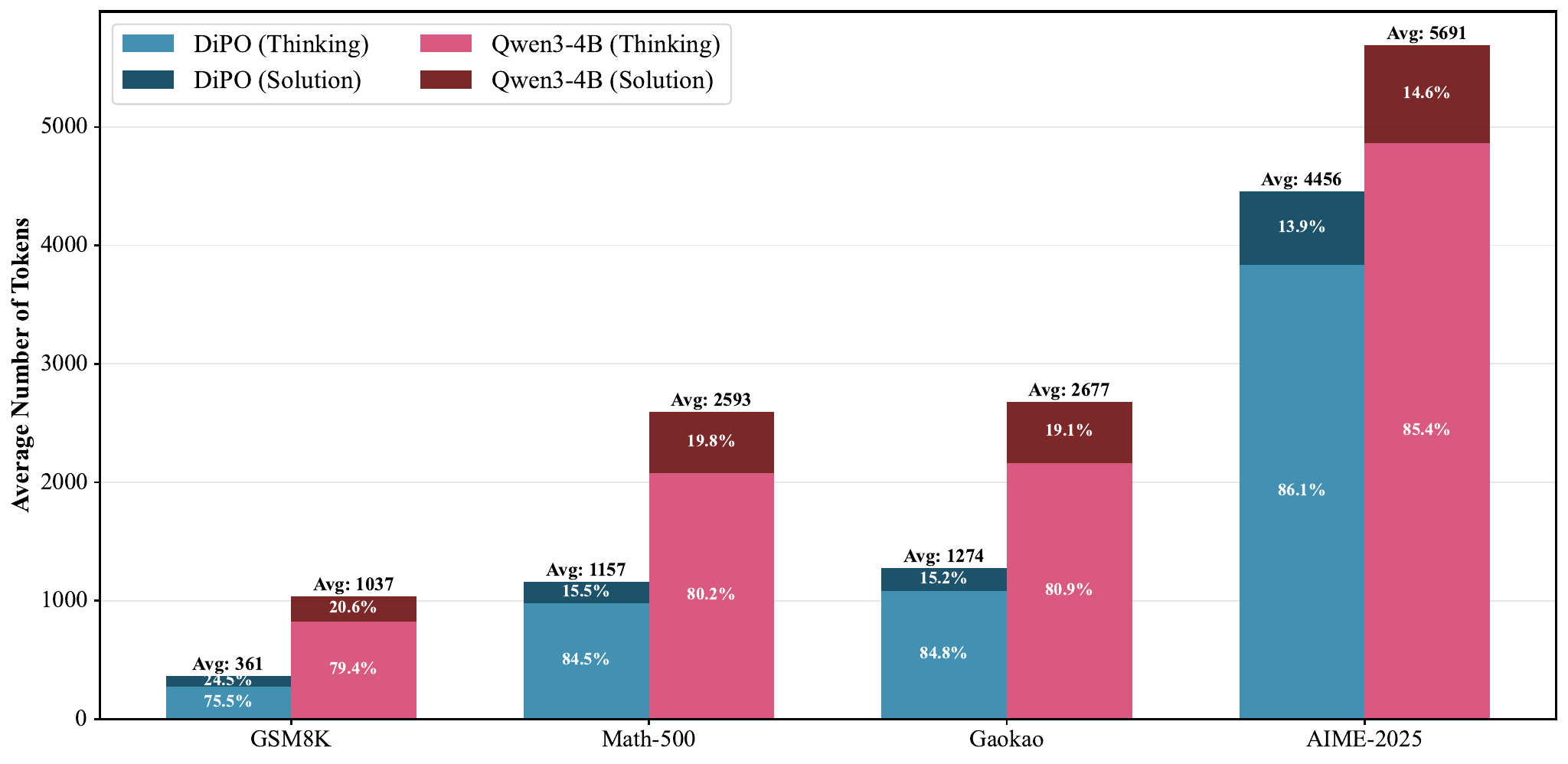}
    \caption{\textbf{Analysis of the Proportion of Thinking Cost.} Average token distribution comparison between the thinking tokens (within the \texttt{<}think\texttt{>} tag) and solution tokens across different datasets. }
    \label{fig:think-Analysis}
\end{figure}

\subsection{Analysis of the Proportion of Thinking Cost}

To investigate the impact of DiPO on the original output structure of the base model, we analyzed the output structures of DiPO and the corresponding base model. Figure \ref{fig:think-Analysis} illustrates the ratio of thinking tokens (those within the \texttt{<}think\texttt{>} tag) to final answer tokens at various generation lengths. We observe that this ratio remains relatively stable across different lengths, with the difference in the proportion of thinking tokens between the two models not exceeding 5\%. This indicates that DiPO effectively compresses the thinking process without introducing redundancy into the final answer, successfully preserving the model's reasoning pattern. As the generation length increases, the last two sets of bars show that the ratio of the thinking chain length stabilizes (86.1\% for DiPO, 85.4\% for the base model). This suggests that, even with an extended reasoning process, the model avoids overly lengthy answers, thus mitigating performance degradation. Consequently, DiPO enhances reasoning efficiency while maintaining the model’s reasoning capability, demonstrating both strong stability and high efficiency.

\subsection{Effect of Difficulty Modeling}

\begin{figure}
    \centering
    \includegraphics[width=1\linewidth]{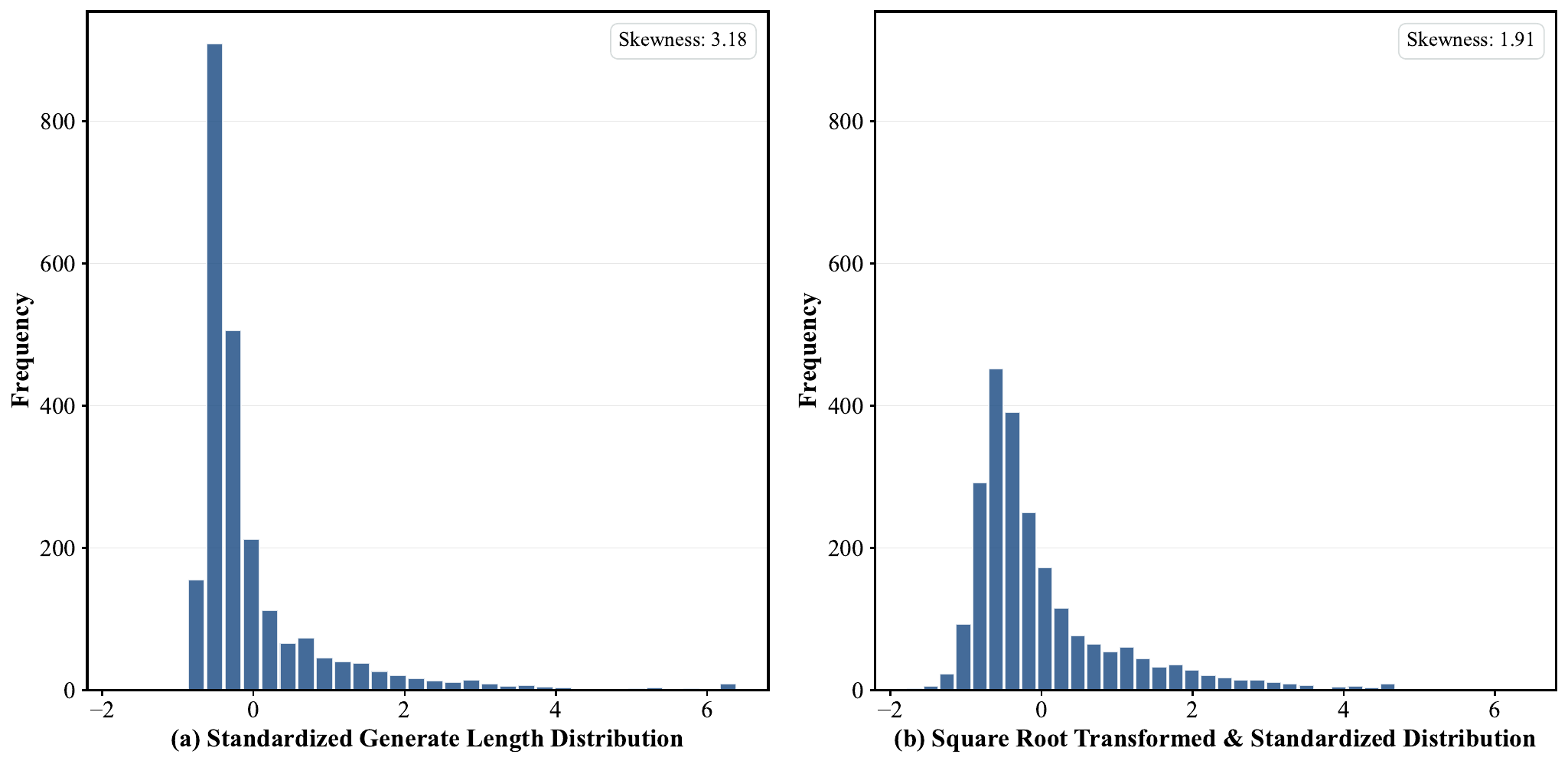}
    \caption{\textbf{Histogram of the Distribution of Difficulty Signals.}}
    \label{fig:diff model}
\end{figure}
In the process of difficulty signal modeling, we employed DeepSeek-V3 to answer all the questions in the training set and recorded the generation length of each question (i.e., the number of tokens in the answer). The original distribution of generation lengths exhibited a significant long-tail characteristic, with a few instances showing exceptionally long generation lengths. This resulted in a highly skewed distribution with a skewness of 3.18. After standardization (mean = 0, variance = 1), as shown in Figure \ref{fig:diff model}-a, this distribution caused extreme samples to dominate the reward function, disrupting training stability. To mitigate this issue, we applied a square root transformation to the generation lengths, effectively compressing the impact of extreme values and reducing the skewness of the distribution to 1.91 (see Figure \ref{fig:diff model}-b). This approach preserves the relative differences between samples while reducing the adverse effects of long-tail data on the training process. As a result, we were able to enhance reasoning efficiency and improve the model’s generalization ability while maintaining training stability.

\subsection{Balance of Effective and Efficient}

\begin{figure}
    \centering
    \includegraphics[width=1\linewidth]{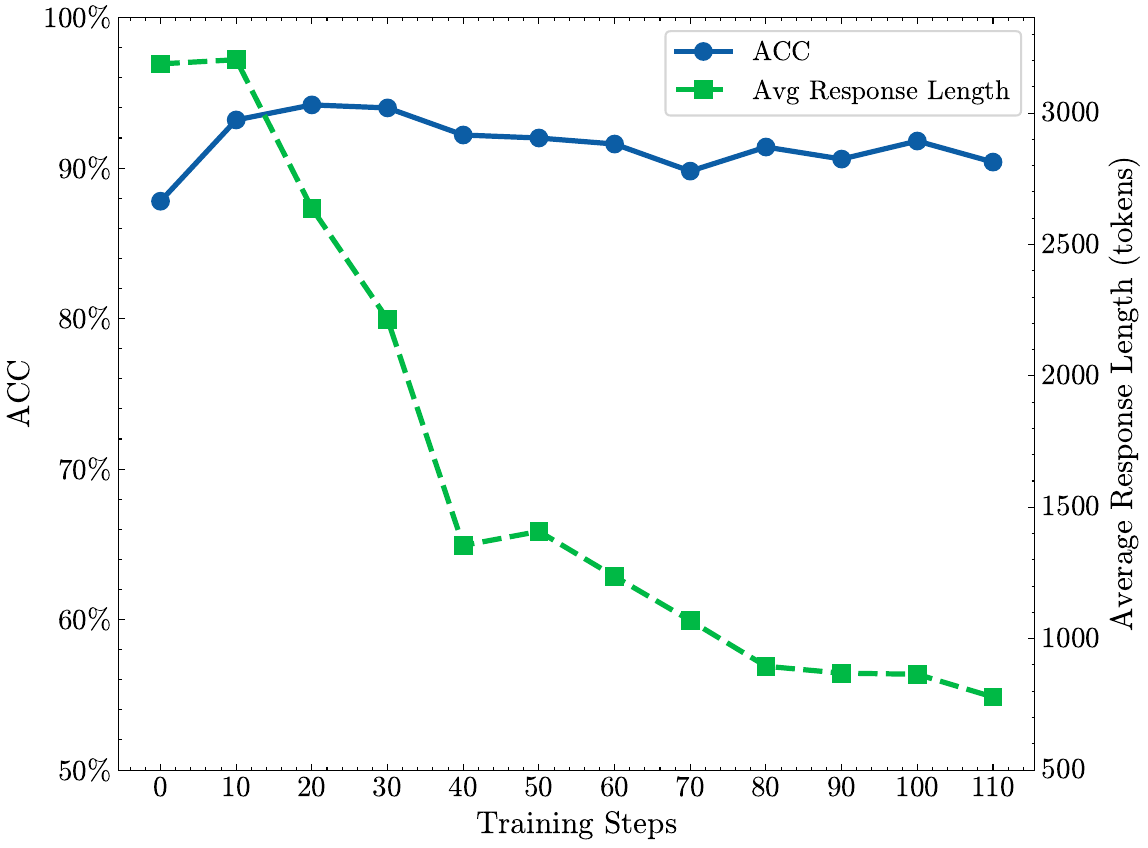}
    \caption{\textbf{Training performance of the Qwen3-4B model on the Math-500 dataset.} The chart shows that as ACC quickly peaks and stabilizes at a high level, the average response length decreases significantly, indicating the model learns to generate answers that are both correct and concise.}
    \label{fig:step analysis}
\end{figure}
We analyzed the trade-off between model Accuracy (ACC) and inference efficiency (Average Response Length) for the Qwen3-4B model on the Math-500 dataset. As depicted in Figure \ref{fig:step analysis}, the model's average response length shows a continuous and significant downward trend during training, with the most rapid decrease occurring in the early stages. Concurrently, the model's ACC rises quickly in the initial phase, peaking at the 20th step before starting to fluctuate. This analysis was designed to identify the optimal training point that maximizes inference efficiency without compromising core reasoning performance by tracking these two key metrics.

Based on a comprehensive analysis of both curves, we ultimately selected the 60th training step as the endpoint, a decision aimed at more aggressively prioritizing inference efficiency. Although the model's ACC peaked at approximately 94.5\% at the 20th step, the response length was still overly long. At the 60th step, the average response length was further compressed to around 1300 tokens—a significant additional efficiency gain compared to the approximately 1800 tokens at the 40th step. While the ACC at this point (around 91.8\%) is slightly lower than the peak, we consider this an acceptable trade-off for achieving maximum compression. More importantly, a sharp drop in performance occurs immediately after 60 steps. Therefore, the 60th step is identified as the optimal point of compression before significant performance degradation, maximizing the reduction of redundant thinking while maintaining a high-accuracy baseline, thus more actively addressing the core objective of this study.

\subsection{Analysis of Reasoning Redundancy via Prompt Constraints}
\label{sec:prompt_analysis}

To rigorously assess whether the reasoning chains optimized by DiPO contain residual redundancy, and to evaluate the method's compatibility with inference-time prompting strategies, we conducted a comparative experiment. We applied two representative prompt-based compression methods, Chain of Draft (CoD) and TALE-EP, directly onto the DiPO-trained Qwen3-4B and Qwen3-8B models. This setup aims to verify whether the model's efficiency can be further improved by external constraints or if DiPO has already converged to the minimal essential reasoning length required for accurate problem-solving.

The quantitative results across GSM8K, Math-500, and GaoKao datasets are summarized in Table~\ref{tab:prompt_compatibility}. The data reveals a distinct trade-off: while incorporating prompt-based constraints (w/ CoD or w/ TALE) successfully reduces the response length and ostensibly improves the information density ratio, it consistently leads to a degradation in accuracy. This phenomenon indicates that the two optimization strategies are not complementary in this context. For instance, on the Math-500 benchmark using the Qwen3-8B model, applying CoD reduces the token count from 1360.1 to 1020.0. However, this 25\% reduction in length comes at a steep cost: the accuracy plummets from 83.40\% to 67.00\%, a substantial decline of 16.4\%. Similarly, on the GaoKao dataset, the TALE-EP method significantly shortens the output but drags the accuracy down from 74.55\% to 60.00\%.

These findings provide strong inverse evidence for the effectiveness of our proposed framework. If DiPO merely produced shorter but still redundant chains, external prompts should be able to compress them further with minimal performance loss. However, the observed sensitivity to forced compression suggests that the reasoning tokens generated by DiPO are functionally necessary. The method has effectively identified and retained the critical cognitive steps required for derivation while discarding ``overthinking" noise. Consequently, enforcing additional brevity via prompts forces the model to skip essential intermediate verification or deduction steps, and thereby leads to under-thinking, which directly impairs the final answer. Furthermore, the larger Qwen3-8B model exhibits higher sensitivity to prompt interference compared to the 4B model, suggesting that as model capacity increases, the internal reasoning structures formed by DiPO become more intricate and resistant to crude external simplification.

\begin{table}[t]
\centering
\setlength{\tabcolsep}{2pt}
\caption{Compatibility analysis of DiPO combined with prompt-based methods. \textbf{Bold} indicates best results. Fitting this into a single column requires significant scaling.}
\label{tab:prompt_compatibility_single}
\resizebox{\columnwidth}{!}{%
\begin{tabular}{lccccccccc}
\toprule
\multirow{2}{*}[-0.5ex]{Method} & \multicolumn{3}{c}{GSM8K} & \multicolumn{3}{c}{Math-500} & \multicolumn{3}{c}{GaoKao} \\
\cmidrule(lr){2-4} \cmidrule(lr){5-7} \cmidrule(lr){8-10}
 & ACC & LEN & Ratio & ACC & LEN & Ratio & ACC & LEN & Ratio \\
\midrule
\multicolumn{10}{l}{\textit{Qwen3-4B}} \\
DiPO    & \textbf{94.31} & 362.5 & 0.582 & \textbf{92.20} & 1354.8 & 0.513 & \textbf{79.74} & 1490.8 & 0.593 \\
w/ CoD  & 92.72 & 198.7 & 0.627 & 82.80 & \textbf{566.4} & \textbf{0.612} & 73.25 & \textbf{623.9} & \textbf{0.693} \\
w/ TALE & 92.65 & \textbf{188.6} & \textbf{0.707} & 86.60 & 586.1 & 0.603 & 70.13 & 644.4 & 0.680 \\
\midrule
\multicolumn{10}{l}{\textit{Qwen3-8B}} \\
DiPO    & \textbf{93.78} & 494.7 & \textbf{0.630} & \textbf{83.40} & 1360.1 & 0.567 & \textbf{74.55} & 1575.2 & 0.595 \\
w/ CoD  & 91.05 & 466.4 & 0.504 & 67.00 & 1020.0 & 0.442 & 55.32 & 1071.9 & 0.478 \\
w/ TALE & 90.52 & \textbf{458.5} & 0.517 & 75.00 & \textbf{697.4} & \textbf{0.593} & 60.00 & \textbf{766.3} & \textbf{0.631} \\
\bottomrule
\end{tabular}%
}
\label{tab:prompt_compatibility}
\end{table}

\subsection{Ablation Study on DiPO Components}
\label{sec:ablation_DiPO}

\begin{figure}[t]
\centering
\includegraphics[width=\columnwidth]{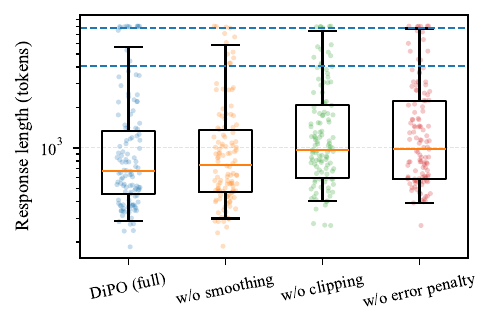}
\caption{Response-length distribution on Math-500 (log scale). Removing error penalty or clipping produces a markedly heavier upper tail, while removing smoothing changes the distribution only slightly.}
\label{fig:ablation_len_boxplot}
\end{figure}

\begin{figure}[t]
\centering
\includegraphics[width=\columnwidth]{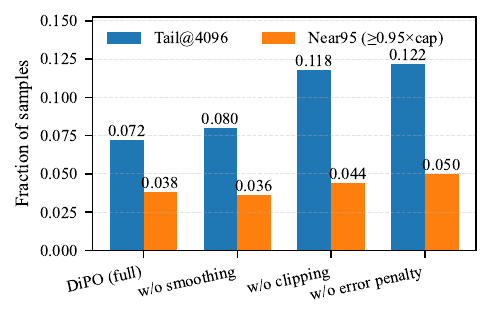}
\caption{Tail probability comparison on Math-500. Tail@4096 measures $\Pr[\mathrm{Len}>4096]$ and Near95 measures $\Pr[\mathrm{Len} \ge 0.95\times8192]$. Error penalty and clipping substantially reduce tail mass.}
\label{fig:ablation_tail_bar}
\end{figure}

We study which components in DiPO drive the accuracy--efficiency trade-off by ablating them on Math-500 (500 problems) under a fixed decoding cap (8192 tokens).
Concretely, we remove one component at a time: (i) length smoothing in the difficulty signal, (ii) an error penalty that down-weights incorrect trajectories, and (iii) clipping that prevents extreme signal values and overly aggressive updates.
Table~\ref{tab:ablation_math500} summarizes overall accuracy (Acc), accuracy on the hardest bucket $(4,5]$ (HardAcc), and response-length statistics.
To go beyond the mean, we additionally visualize the length distribution with a log-scale boxplot (Figure~\ref{fig:ablation_len_boxplot}) and report tail probabilities (Figure~\ref{fig:ablation_tail_bar}).

\begin{table}[t]
\centering
\setlength{\tabcolsep}{3.2pt}
\renewcommand{\arraystretch}{1.08}
\caption{Ablation on Math-500 (500 problems, cap=8192). HardAcc is accuracy on the hardest bucket $(4,5]$. Tail@4096 is $\Pr[\mathrm{Len} > 4096]$, and Near95 is $\Pr[\mathrm{Len} \ge 0.95\times8192]$.}
\label{tab:ablation_math500}
\resizebox{\columnwidth}{!}{%
\begin{tabular}{lccccc}
\toprule
\multirow{2}{*}[-0.5ex]{Method} & \multicolumn{2}{c}{Accuracy} & \multicolumn{3}{c}{Length / Tail (tok)} \\
\cmidrule(lr){2-3} \cmidrule(lr){4-6}
 & Acc$\uparrow$ & HardAcc$\uparrow$ & Len$\downarrow$ & Tail@4096$\downarrow$ & Near95$\downarrow$ \\
\midrule
DiPO (full)        & 0.922 & 0.858 & 1347.5 & 0.072 & 0.038 \\
w/o smoothing      & 0.912 & 0.851 & 1399.8 & 0.080 & 0.036 \\
w/o error penalty  & 0.916 & 0.821 & 1860.5 & 0.122 & 0.050 \\
w/o clipping       & 0.908 & 0.853 & 1812.8 & 0.118 & 0.044 \\
\bottomrule
\end{tabular}%
}
\end{table}

\noindent\textbf{Length smoothing.}
Removing smoothing causes a small but consistent regression in accuracy (Acc: 0.922$\rightarrow$0.912; HardAcc: 0.858$\rightarrow$0.851) and slightly increases the mean length (1347.5$\rightarrow$1399.8).
The boxplot shows nearly unchanged central mass, and the tail metrics shift only marginally (Tail@4096: 0.072$\rightarrow$0.080; Near95: 0.038$\rightarrow$0.036).
Overall, smoothing provides a mild but consistent regularization effect, improving the stability of length control without materially changing the overall distribution.

\noindent\textbf{Error penalty.}
This component is critical for mitigating unproductive long generations on hard instances.
Without the error penalty, responses become much more heavy-tailed (Len=1860.5; Tail@4096=0.122; Near95=0.050), and the upper tail in the boxplot increases noticeably.
HardAcc drops substantially (0.858$\rightarrow$0.821) while overall Acc changes only slightly, indicating that on the hardest subset the model more often allocates additional tokens to ineffective trajectories instead of reaching correct solutions.

\noindent\textbf{Clipping.}
Clipping serves as a safeguard against tail bursts and unstable updates.
Ablating clipping increases the mean length (Len=1812.8) and consistently worsens tail behavior (Tail@4096=0.118; Near95=0.044), consistent with the heavier upper tail in Figure~\ref{fig:ablation_len_boxplot}.
The weakened length control suggests that clipping prevents extreme updates that lead to excessively long generations.

\noindent\textbf{Summary.}
Across ablations, the error penalty and clipping account for most of DiPO's length-control benefits.
Removing either component noticeably increases long-tail probability and reduces accuracy on the hardest bucket, while overall accuracy changes comparatively little.
Length smoothing yields smaller but consistent gains, mainly improving stability of the length distribution.
Overall, DiPO's efficiency stems from suppressing low-quality trajectories on difficult instances and avoiding unstable optimization dynamics, rather than from directly enforcing shorter generations.

\section{Conclusion}
This paper proposes DiPO, a post-training framework based on reinforcement learning to mitigate the frequent overthinking issue of LRMs. DiPO aims to train the model to perceive task difficulty and adaptively adjust reasoning depth. The difficulty modeling method based on model self-reasoning can formalize task complexity while eliminating dependence on manual annotation. The reward function that integrates difficulty signals achieves compatibility between performance and efficiency of reasoning. We conduct extensive comparative experiments on seven public datasets, including in-domain and out-of-domain. Experimental results show that DiPO significantly improves inference efficiency compared to the baselines without compromising performance. This study provides new perspectives and unique insights for future research.


\bibliographystyle{IEEEtran}

\bibliography{overthink}

\end{document}